\pdfoutput=1

\documentclass[11pt]{article}

\usepackage{emnlp2021}

\usepackage{booktabs}
\usepackage{balance}
\usepackage{times}
\usepackage{latexsym}

\usepackage[T1]{fontenc}

\usepackage[utf8]{inputenc}

\usepackage{microtype}

\usepackage{graphicx}
\usepackage{mathtools}
\usepackage{amsmath,amsfonts,amsthm,bm}

\def\sR{{\mathbb{R}}}

\def\gE{{\mathcal{E}}}

\def\gK{{\mathcal{K}}}
\def\gL{{\mathcal{L}}}

\def\gQ{{\mathcal{Q}}}
\def\gR{{\mathcal{R}}}

\usepackage{xcolor}
\usepackage{multirow}

\usepackage{tabularx,colortbl}
\usepackage[most]{tcolorbox}
\definecolor{Gray}{HTML}{ABD5FF}  %
\definecolor{lightgray}{rgb}{.9,.9,.9}
\definecolor{darkgray}{rgb}{.4,.4,.4}
\definecolor{purple}{rgb}{0.65, 0.12, 0.82}

\definecolor{lightgray}{rgb}{.9,.9,.9}
\definecolor{darkgray}{rgb}{.4,.4,.4}
\definecolor{purple}{rgb}{0.65, 0.12, 0.82}

\newcommand{\ours}{Grounded Graph Decoding}

\title{\ours{} Improves Compositional\\Generalization in Question Answering}

\author{
    Yu Gai\textsuperscript{\textnormal{{\large\textasteriskcentered}}} \and Paras Jain\textsuperscript{\textnormal{{\large\textasteriskcentered}}} \and Wendi Zhang \and \\{\bf Joseph E. Gonzalez} \and {\bf Dawn Song} \and {\bf Ion Stoica} \\
    University of California, Berkeley \\
    \texttt{\{yu\_gai, paras\_jain\}@berkeley.edu} \\
    \textsuperscript{\large\textasteriskcentered}equal contribution \\
}

\begin{document}

\maketitle

\begin{abstract}
Question answering models struggle to generalize to novel compositions of training patterns, such to longer sequences or more complex test structures. Current end-to-end models learn a flat input embedding which can lose input syntax context. Prior approaches improve generalization by learning permutation invariant models, but these methods do not scale to more complex train-test splits. We propose Grounded Graph Decoding, a method to improve compositional generalization of language representations by grounding structured predictions with an attention mechanism. Grounding enables the model to retain syntax information from the input in thereby significantly improving generalization over complex inputs. By predicting a structured graph containing conjunctions of query clauses, we learn a group invariant representation without making assumptions on the target domain. Our model significantly outperforms state-of-the-art baselines on the Compositional Freebase Questions (CFQ) dataset, a challenging benchmark for compositional generalization in question answering. Moreover, we effectively solve the MCD1 split with 98\% accuracy. All source is available at \url{https://github.com/gaiyu0/cfq}.
\end{abstract}

\section{Introduction}
Can neural networks ``\textit{make infinite use of finite means}" with language~\cite{chomsky2002syntactic}? The ability of humans to reason compositionally enables us to form novel complex compound sentences by combining constituent concepts. Toki Pona~\cite{lang2014tokipona} is an engineered language with only 120 words but can express a wide variety of concepts through composition.

Compositionality specifically refers to the phenomenon that the meaning of an expression is given by combining the meanings of its parts~\cite{montague1970universal}.
For example, after understanding the questions ``Who directed Inception?” and ``Did Christopher Nolan produce Goldfinger?”, one can understand ``Who produced Inception?".

\begin{figure}[t]
    \includegraphics[scale=0.36]{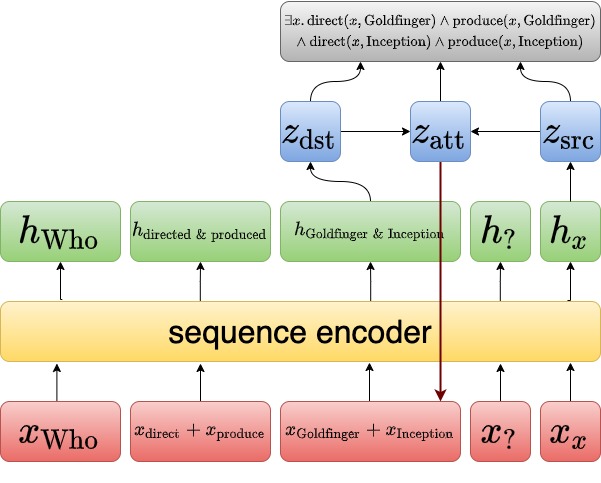}
    \caption{
        An illustration of \ours{} with the question ``Who directed and produced Goldfinger and Inception?".
        The predicates ``direct(ed)" and ``produce(d)", and the entities ``Goldfinger" and ``Inception" are grouped together by the sequence encoder.
        Black lines represent linear transforms between latent spaces, while the red line represents the attention that grounds the prediction.
    }
    \label{fig:arch}
\end{figure}

However, it's unclear whether neural networks truly reason compositionally; for example, \citet{min2019compositional} argue that compositional reasoning is not necessary to answer complex composite questions.
To benchmark the performance of neural models under compositional generalization, \citet{lake2018generalization} propose the SCAN dataset containing sequences of instructions. SCAN uses length of input as a proxy for compositional complexity. However, SCAN's length-based splits are now considered solved~\cite{chen2020compositional}.

As a more realistic benchmark, \citet{keysers2019measuring} introduced the Compositional Freebase Questions (CFQ) dataset. CFQ is a challenging knowledge graph question answering task with adversarial splits (MCD1, MCD2 and MCD3) that maximize compositional divergence between train and test sets. While a simple LSTM model achieves near-perfect accuracy on an i.i.d. split, large models such as T5-11B~\cite{raffel2019exploring} performs catastrophically poorly on the MCD splits.

We hypothesize current models perform poorly on the CFQ task as they fail to take semantic and syntactic structures into account. We propose \ours{} to incorporate these structures with graph decoding and grounding. The encoder often fails to match common syntactic patterns from the input text question. We modify the decoder with attention over the input sequence to enable referencing specific input syntactic structures. Moreover, we find the semantics of knowledge graph queries (SPARQL) is a \textit{conjunctive query graph}. Sequence decoders fail to represent this structure well. We integrate semantic structure by decoding output predictions as a graph.

Our work makes the following contributions:
\begin{itemize}
    \item We ground the decoder using the input text in order to improve understanding of questions with novel syntactic structures.
    \item We leverage a conjunctive query graph decoder to enable generation of novel complex SPARQL queries.
    \item On the challenging Compositional Freebase Questions (CFQ) benchmark, \ours{} effectively solves the MCD1 split while achieving state-of-the art results on the remaining two splits.
\end{itemize}

\section{Preliminaries}\label{sec:prelim}
We formally present an overview of compositional generalization in deep learning to motivate \ours{}.

\subsection{Compositional generalization and its benchmarks}
Training models that achieve compositional generalization would be an important advancement for both practical systems and for furthering our understanding of intelligence more broadly as compositional intelligence is a key characteristic of the human mind.
Moreover, compositional generalization would improve the sample efficiency of models, as argued by~\citet{lake2018generalization}.
This would improve answers for rare questions while accelerating learning by decomposing the combinatorial nature of language.

Benchmarks for compositional generalization measure how well neural models trained on one set of structures generalize to an unseen test set with novel structures.
\citet{lake2018generalization} experimented with several heuristics for splitting seq2seq datasets, such as splitting by sequence length, and found that some of them posed significant challenge for state-of-the-art sequence-to-sequence (seq2seq) models to generalize.

\citet{keysers2019measuring} take a systematic approach based on the distinction between atoms and compounds.
For example, in the questions ``Who directed Inception?'' and ``Did Christopher Nolan produce Goldfinger?'', the atoms are the primitive elements that form these questions.
These elements include the predicates ``direct(ed)” and ``produce(d)'', the question patterns ``Who [predicate] [entity]'' and ``Did [entity1] [predicate] [entity2]'', and the entities “Inception”, “Christopher Nolan”, etc.
Compounds are intuitively combinations of atoms, such as ``Who directed [entity]?'', which is the combination of the predicate ``direct(ed)" and the question template ``Who [predicate] [entity]?'', etc.
\citet{keysers2019measuring} proposes the following objective for partitioning the dataset:
(a) minimum atom divergence, which ensures atoms that occur in test sets also occur in training sets and (b) maximum compound divergence, which maximizes the number of compounds in test sets that are not present in training sets.

Intuitively, in order to succeed on such splits, a model has to both learn the meaning of atoms, and learn the rules that combine the atoms.
A reasonable question to appear in the test set of a MCD split, if the two questions above occur in the training set of the split, can be ``Who produced Goldfinger?", which consists of only known atoms, but entirely different compositions.

\subsection{Conjunctive queries and knowledge graph question answering (KG-QA)}

Many factual questions can be answered by executing conjunctive queries against knowledge graphs (KG).
Formally, given a KG $\gK = (\gE, \gR)$ that consists of a set of entities $\gE$ and a set of relation triples $\gR$, a conjunctive query \[
    \exists x_1, \ldots, x_n. r_1 (s_1, o_1) \land \ldots \land r_m (s_m, o_m)
\] against $\gK$ is the conjunction of predicates $r_1 (s_1, o_1), \ldots, r_m (s_m, o_m)$.
The subject $s_i$ and object $o_i$ of each predicate can be either an entity in $\gE$, or one of the variables $x_1, \ldots, x_n$, and predicate $r_i (s_i, o_i)$ is true if and only if relation $r_i$ holds between $s_i$ and $o_i$ in $\gK$.
For example, to answer the question ``Who directed and produced Inception?", we can use the conjunctive query ``$\exists x. \text{direct} (x, \text{Inception}) \land \text{produce} (x, \text{Inception})$".
Although general queries, such as lambda dependency-based compositional semantics~\citep{liang2013lambda} are also popular in the KG-QA literature, both SCAN and CFQ find that parsing questions into conjunctive queries alone presents a nontrivial challenge to compositional generalization despite their simplicity.

Compared to other question answering settings, such as reading comprehension question answering (RC-QA), KG-QA is exclusively concerned with question comprehension, making it an ideal unit test for QA models.
We consider the \emph{strongly supervised} setting of KG-QA, where each question for training is paired with a query that, when executed against a KG, yields the question's answer.
Although this is less challenging than the \emph{weakly supervised} setting, where only answers to questions are available, both \citep{keysers2019measuring} and \citep{furrer2020compositional} show that this setting already requires a level of compositional generalization not yet possessed by existing systems.

\subsection{Compositionality and compositional language representation}
Compositionality refers to the phenomenon that the meaning of an expression is given by combining the meanings of its parts \citep{montague1970universal}.
Specifically, for an expression $z$ composed of subexpressions $x$ and $y$, its semantics $\tau (z)$ is given by \[
    \tau (z) = \tau (x \oplus y) = \tau (x) \odot \tau (y)
\] where $\oplus$ is a syntactic composition operator, $\odot$ is an semantic composition operator, and the semantics $\tau (x)$ and $\tau (y)$ of $x$ and $y$ are given recursively by the same rule.

For example, the meaning of the phrase ``direct and produce" is given by the conjunction of the words ``direct" and ``produce".
In this case the syntactic composition operator $\oplus$ simply takes two words and adds an ``and" in between, while the semantic composition operator $\odot$ yields the conjunction of the two predicates.
This phrase can be further composed with other phrases to express more complex meanings.
Various formalisms exist for this process, such as the Compositional Categorial Grammar \citep{zettlemoyer2012learning}.

The key question is how to model semantics in vector spaces.
Motivated by \citet{montague1970universal}, \citet{andreas2019measuring} defines a neural model as compositional if it is a homomorphism from syntax trees to vector representations, that is, for an expression $z = x \oplus y$, its representation $\theta (z)$ is given by: \[
    \theta (z) = \theta (x \oplus y) = \theta (x) \odot \theta (y)
\]

This formulation should be simple to implement if both the syntactic and semantic structure of a language is known.
However, a key challenge to building compositional natural language representation is the lack of both syntactic and semantic structure.
First, natural languages only loosely follow the compositionality principle.
Especially, natural languages are not always context-free, which is implicitly assumed in this formulation of compositionality.
The semantics $\tau (x)$ can depend on only the phrase $x$ only and not on phrase $y$.
Second, the syntactic structure is not always known.

Despite the excellent performance of constituency parsers on benchmarks, they may generalize poorly to compositionally complex questions (see Section~\ref{sec:experiments} for more discussion).
Despite substantial progress in the NLP community to learn vector representations of semantics, how to learn compositionally generalizable vector representations of semantics without much knowledge of compositional structures remains challenging.

\section{Methodology: \ours{}}\label{sec:model}

In the following sections, we first describe how to decode conjunctive query graphs from natural language questions, then how to incorporate syntactic compositions in graph decoding, and finally introduce a grounding mechanism by explicitly conditioning graph decoding on syntactic compositions.

\subsection{Graph decoding}\label{sec:seq2gr}
A conjunctive query can be naturally represented as a directed graph by representing subjects and objects in the query as nodes, and relational predicates in the query as directed edges.
For example, the conjunctive query \[
    \exists x_1. \text{direct} (x_1, \text{Inception}) \land \text{produce} (x_1, \text{Inception})
\] can be represented as a directed graph with nodes $x_1$ and ``Inception", and edges \[
    x_1 \xrightarrow{\text{direct}} \text{Inception} \qquad
    x_1 \xrightarrow{\text{direct}} \text{Goldfinger}
\] where the type of an edge is the relation in the corresponding relational predicate.
The key benefit of the directed graph representation is permutation invariance.
Directed graphs are invariant to ordering of edges as conjunctive queries are invariant to ordering of relational predicates.

Similar to the graph decoder architecture proposed by \citet{kipf2016variational}, our graph decoder first generates embeddings for nodes, and then predicts edges in graphs using the node embeddings.
For simplicity, we assume that questions are tokenized into words, and entities in questions have been found and anonymized~\citep{finegan2018improving}.
The embedding of an entity node is simply the contextual embedding of its mentions in the question.
If an entity is mentioned multiple times in the question, its node embedding is the sum of the contextual embeddings of all its mentions.
Given a question $q$ consisting of tokens $q_1, \ldots, q_l$, we concatenate the question with the list of variables $x_1, \ldots, x_n$ into a new sequence \[
    q_1, \ldots, q_l, \text{[SEP]}, x_1, \ldots, x_n
\] and input this sequence to a sequence encoder, such as an LSTM~\citep{hochreiter1997long}.
The embedding of the variable node $x_i$ is the contextual embedding of the token $x_i$.
Intuitively, these contextual node embeddings capture relations between both entity and variable nodes.
We denote the embedding for node $v$ as $h_v$.

Given the node embeddings, the probability that an edge $s \xrightarrow{r} o$ exists between subject $s$ and object $o$ is modeled as \begin{align}
    P(s \xrightarrow{r} o | q) = \sigma \left(w_r^T [h_s, h_o] \right)
\end{align} where $\sigma$ denotes the sigmoid function, $w_r^T \in \sR^{2 d}$ is the weight vector specific to relation $r$, $h_s$ and $h_o$ are the node embeddings of subject $s$ and object $o$ respectively, and $[\cdot, \cdot]$ denotes vector concatenation.
Multiple relations may hold between an entity pair.

The model is trained to maximize the conditional log-likelihood of all conjunctive query graphs in training set $\gQ$: \begin{align}
    \gL
    & = \sum_{q \in \gQ} \log P(\tau (q) | q) \nonumber \\
    & = \sum_{q \in \gQ} \sum_{s \xrightarrow{r} o \in \gE (\tau (q))} \log P (s \xrightarrow{r} o | q) \nonumber \\
    & \qquad + \sum_{s \xrightarrow{r} o \not \in \gE (\tau (q))} \log (1 - P (s \xrightarrow{r} o | q)) \label{jll}
\end{align} where $\tau (q)$ denotes the conjunctive query graph for question $q$, and $\gE (\tau (q))$ denotes the edges in $\tau (q)$.

\subsection{Incorporating syntactic compositions}\label{sec:cseq2gr}

Although our graph decoder incorporates the compositional structure of conjunctive queries, it still cannot capture syntactic compositions in natural language questions due to the lack of compositionality in the sequence encoder.
A solution to the problem is to embed syntactic structures in the sequence encoder.
In general, however, these structures can be hard to identify.

We manually evaluated parses returned by the Stanford constituency parser~\cite{bauer_2014} given questions in the CFQ dataset, and observed a high error rate, possibly due to complex compositional structures in the questions.
As a pilot study, instead of incorporating syntactic compositions exhaustively, we only consider a simple syntactic composition ``A and B", where ``A" and ``B" share the same part-of-speech.
This composition can be reliably identified using part-of-speech (PoS) tagging.
Following the suggestion of \citet{andreas2019measuring}, the embeddings $\theta$ that model semantic compositions should satisfy \[
    \theta (\text{``A and B"}) = \theta (\text{``A"}) \odot \theta (\text{``B"})
\]
A key criterion for choosing the semantic composition operator $\odot$ is \emph{permutation invariance}, which is evident from the fact that ``A and B" has the same meaning as ``B and A".
We model the semantics of both ``A" and ``B" using learnable vector embeddings, and model the semantic composition using vector addition.
This model can be easily extended to the case where both ``A" and ``B" are phrases, in which case both $\theta (\text{``A"})$ and $\theta (\text{``B"})$ can be embeddings generated by sequence encoders, and $\odot$ can be any learnable composition operator.
However, for simplicity, we only consider the case that ``A" and ``B" are single words.
To implement this idea, instead of inputting \[
    q_1, \ldots, q_l, \text{[SEP]}, x_1, \ldots, x_n
\] to the sequence encoder, we now merge the phrases like ``A and B" into a single token, which we call a group, and input the groups to a sequence encoder \[
    g_1, \ldots, g_k, \text{[SEP]}, x_1, \ldots, x_n
\]
The embedding $x_g$ of a group $g$ is given by \[
    \theta (g) = \sum_{w \in g} \theta (w)
\] where $\theta (w)$ is the embedding of word $w$.
For example, the embedding of ``A and B" is $\theta (\text{A}) + \theta (\text{B})$.
The sum can be replaced by learnable composition operators to model more complex semantic compositions.

When an entity is mentioned in a group, we use the contextual embedding of the group as its node embedding.
The probability for the conjunctive query graph to contain a predicate $s \xrightarrow{r} o$ is now modeled as \begin{equation}
    P(s \xrightarrow{r} o | q)
    = \sigma \left(w_r^T [g_s, g_o] \right) \label{prob}
\end{equation} and all parameters are still learned by maximizing the joint log-likelihood in Eqn.~\ref{jll}.

\subsection{Grounded graph decoding}

Curiously, we find in preliminary experiments that although incorporating syntactic compositions in the sequence encoder improves compositional generalization, the incorporated syntactic compositions are not always reflected in the model's outputs.
For example, given the question ``Who directed and produced Inception?", the model may output only the predicate $\text{direct} (x, \text{Inception})$ but not the other predicate $\text{produce} (x, \text{Inception})$, despite ``produced" being grouped with ``directed" by the sequence encoder.

The phenomenon indicates that syntactic compositions are insufficiently preserved, and the model actually fails to learn the correspondence between syntactic and semantic compositions.
To encourage the model to learn the correspondence, we propose to augment the graph decoder with a mechanism that enables it to ground its outputs in syntactic compositions.
Specifically, we add a grounded embedding $z_{s, o}$ to Eqn.~\ref{prob}:
\[
    P(s \xrightarrow{r} o | q) = \sigma (w_r^T [h_s, h_o, z_{s, o}])
\]
The grounded embedding $z_{s, o}$ is given by an weighted average over syntax compositions.
Mathematically, \[
    z_{s, o}
    = \sum_{k = 1}^l \alpha_{s, o}^{(k)} \nu_k
    = \sum_{k = 1}^l \alpha_{s, o}^{(k)} \sum_{w \in g_k} x_w
\]
The attention weight $\alpha_{s, o}^{(k)}$, which quantifies the relevance of the $k$-th group to the subject-object pair $(s, o)$, is given by \[
    \alpha_{s, o}^{(k)}
    = \frac{\exp (a_{s, o}^{(k)} / \sqrt{d})}{\sum_{j = 1}^l \exp (a_{s, o}^{(j)} / \sqrt{d})}
\] where following \citet{vaswani2017attention}, we set the temperature of the softmax function to $\sqrt{d}$, square root of the latent space dimension.
The unnormalized attention score $a_{s, o}^{(k)}$ is given by an inner product between the query $q_{s, o}$ and the key $\kappa_k$ \[
    a_{s, o}^{(k)} = q_{s, o}^T \kappa_k
\]
Both the query $q_{s, o}$ and the key $\kappa_k$ are given by linear transforms of their contextual embedding \[
    q_{s, o} = Q [h_s, h_o] \qquad
    \kappa_k = K h_k
\]
All parameters in the model are still learned by maximizing the joint log-likelihood in Eqn.~\ref{jll}.
We append a special ``NIL" token to the end of each question, so that the graph decoder can attend to this token when no relation exists between a pair of subject and object.
See Fig.~\ref{fig:arch} for an illustration of the architecture.
As this embedding is specific to node pairs, the model is able to ground different edges in different syntactic compositions.

\section{Experiments}\label{sec:experiments}

\begin{table*}[ht]
\centering
\resizebox{\textwidth}{!}{ %
\begin{tabular}{@{}llllll@{}}
\toprule
\multicolumn{1}{c}{\multirow{2}{*}{\textbf{Method}}} &
  \multicolumn{1}{c}{\multirow{2}{*}{\textbf{\# Params.}}} &
  \multicolumn{3}{c}{\textbf{Accuracy per-split}} \\
\multicolumn{1}{c}{} &
  \multicolumn{1}{c}{} &
  \multicolumn{1}{c}{\textit{MCD1}} &
  \multicolumn{1}{c}{\textit{MCD2}} &
  \multicolumn{1}{c}{\textit{MCD3}} \\ \midrule
LSTM w/ attention~\cite{keysers2019measuring} & & 28.9 $\pm$ 1.8\% & 5.0 $\pm$ 0.8\%  & 10.8 $\pm$ 0.6\% \\
Transformer~\cite{keysers2019measuring} & & 34.9 $\pm$ 1.1\% & 8.2 $\pm$ 0.3\%  & 10.6 $\pm$ 1.1\% \\
Universal Transformer~\cite{keysers2019measuring} & & 37.4 $\pm$ 2.2\% & 8.1 $\pm$ 1.6\%  & 11.3 $\pm$ 0.3\% \\
Evolved Transformer~\cite{keysers2019measuring} & & 42.4 $\pm$ 1.0\% & 9.3 $\pm$ 0.8\%  & 10.8 $\pm$ 0.2\% \\
T5-base~\cite{furrer2020compositional} & 220M & 57.6 $\pm$ 1.4\% & 19.5 $\pm$ 1.0\% & 16.6 $\pm$ 1.5\% \\
T5-large~\cite{furrer2020compositional} & 770M & 63.3 $\pm$ 0.6\% & 22.2 $\pm$ 1.5\% & 18.8 $\pm$ 2.6\% \\
T5-11B~\cite{furrer2020compositional} & 11000M & 61.4 $\pm$ 4.8\% & 30.1 $\pm$ 2.2\% & 31.2 $\pm$ 5.7\% \\
T5-11B (modified)~\cite{furrer2020compositional} & 11000M & 61.6 $\pm$ 12.4\% & 31.3 $\pm$ 12.8\%& 33.3 $\pm$ 2.3\% \\
\midrule
\cellcolor{Gray}\ours{} & \cellcolor{Gray}0.3M  & \cellcolor{Gray}\textbf{97.9 $\pm$ 0.2\%} & \cellcolor{Gray}\textbf{47.1 $\pm$ 10.4\%} & \cellcolor{Gray}\textbf{50.8 $\pm$ 17.2\%} \\
\bottomrule
\end{tabular}
} %
\caption{
    \textbf{CFQ evaluation without tuning on development sets} \ours{} achieves significantly higher performance than state-of-the-art seq2seq baselines across all MCD splits.
    As the MCD1 accuracy of \ours{} is withing the range of baselines trained on the random split ($\sim98\%$), we consider MCD1 to be solved.
}
\label{tab:not_on_dev}
\end{table*}

\begin{table*}[]
\centering
\begin{tabular}{@{}llllll@{}}
\toprule
\multicolumn{1}{c}{\multirow{2}{*}{\textbf{Method}}} &
  \multicolumn{1}{c}{\multirow{2}{*}{\textbf{\# Params.}}} &
  \multicolumn{3}{c}{\textbf{Accuracy per-split}} \\
\multicolumn{1}{c}{} &
  \multicolumn{1}{c}{} &
  \multicolumn{1}{c}{\textit{MCD1}} &
  \multicolumn{1}{c}{\textit{MCD2}} &
  \multicolumn{1}{c}{\textit{MCD3}} \\ \midrule
Hierarchical Poset Decoding~\cite{guo2020hierarchical} & & 79.6\% & 59.6\% & 67.8\% \\ 
CBR-KGQA \citep{das2021case} & & 87.9\% & 61.3\% & 60.6\% \\
T5-3B \cite{herzig2021unlocking} & 3000M & 65.0\% & 41.0\% & 42.6\% \\
LIR + RIR (T5-3B) \cite{herzig2021unlocking} & 3000M & 88.4\% & \textbf{85.3\%} & \textbf{77.9\%} \\
\midrule
\cellcolor{Gray}\ours{} & \cellcolor{Gray}0.3M & \cellcolor{Gray}\textbf{98.6\%} & \cellcolor{Gray}67.9\% & \cellcolor{Gray}77.4\% \\
\bottomrule
\end{tabular}
\caption{
    \textbf{CFQ evaluation with tuning on development sets}
    \citet{keysers2019measuring} discourage tuning models with development sets in MCD splits as it compromises the divergence split.
    We therefore report separately in Table~\ref{tab:on_dev} results obtained with tuning on development sets.
    Results without tuning on development sets are not reported in \citet{guo2020hierarchical}, \citet{herzig2021unlocking}, and \citet{das2021case}.
}
\label{tab:on_dev}
\end{table*}

\begin{table*}[]
\centering
\begin{tabular}{@{}lcccc@{}}
\toprule
\multicolumn{1}{c}{\textbf{Method}} & \multicolumn{1}{c}{\textbf{MCD1}} & \multicolumn{1}{c}{\textbf{MCD2}} & \multicolumn{1}{c}{\textbf{MCD3}} \\ \midrule
Poset Decoding~\cite{guo2020hierarchical} & 21.3\% & 6.4\% & 10.1\% \\ 
\midrule
\cellcolor{Gray}Grounded Graph Decoding & \cellcolor{Gray}\textbf{98.6\%} & \cellcolor{Gray}\textbf{67.9\%} & \cellcolor{Gray}\textbf{77.4\%} \\
~~~~~~Syntax-aware graph decoding & {76.0\%} & {29.0\%} & {32.7\%} \\
~~~~~~Graph decoding only & {59.1\%} & {25.7\%} & {20.4\%} \\
\bottomrule
\end{tabular}
\caption{
    Ablations of \ours{}.
    Both grounding and grouping significantly improve accuracy on all splits.
    We also compare our bare graph decoder with poset decoding to show the advantage of the directed graph representation.
}
\label{tab:ablations}
\end{table*}

We evaluate \ours{} to understand how grounding as well as graph decoding improve semantic and syntactic understanding.

\subsection{Dataset}

We evaluate our model using the Compositional Freebase Queries (CFQ) dataset \citep{keysers2019measuring}, a semantic parsing dataset consisting of approximately 240k natural language questions paired with conjunctive queries.
Compared to other semantic parsing datasets, the CFQ dataset features richer question patterns, making it an ideal benchmark for compositional generalization (see \citet{keysers2019measuring} for a quantitative comparison).

The CFQ dataset consists of three Maximum-Compound Divergence (MCD) splits (MCD1, MCD2, and MCD3), constructed by running a greedy algorithm with different initializations that maximize compound divergence between training and test sets, while keeping atom divergence between them minimum (see Section~\ref{sec:prelim} for more details about atom and compound divergence).

\subsection{Baselines}
We compare \ours{} with various baselines established by \citet{keysers2019measuring} and \citet{furrer2020compositional}, as well as HPD \citep{guo2020hierarchical}, CBR-KBQA \citep{das2021case}, as well as the fine-tuning scheme by \citet{herzig2021unlocking}.

The baseline seq2seq models include LSTM \citep{hochreiter1997long} with attention \citep{bahdanau2014neural}, Transformer \cite{vaswani2017attention}, Universal Transformer \citep{dehghani2018universal}, Evolved Transformer \citep{so2019evolved}, and T5 \citep{raffel2019exploring}, all of which, except "T5-11B (modified)", output conjunctive queries as sequences of tokens.
``T5-11B (modified)" refers to a scheme proposed by \citet{furrer2020compositional} in which T5 models only need to output conjunctive queries that are aligned with syntactic structures in questions.
All seq2seq baselines are reported in \citet{keysers2019measuring} and \citet{furrer2020compositional}.

Hierarchical Poset Decoding (HPD) follows the encoder-decoder architecture of seq2seq models, but instead of outputting conjunctive queries as sequences of tokens, it outputs them as posets.
Compared to seq2seq models, this has the advantage that posets are partially invariant to permutation of relational predicates, although decoding actually can be made fully permutation invariant using the graph representation described in Section~\ref{sec:model}.

Concurrent work, CBR-KBQA~\cite{das2021case}, is a semi-parametric scheme in which a parametric model parses questions into queries at test by combining relevant queries in training sets.

Concurrent work from \citep{herzig2021unlocking} proposes to fine-tune pre-trained models with not only questions and queries but also intermediate query representations that align with question structures.
(The reported results are obtained using a T5 model with 3 billion parameters).
Although this scheme is model-agnostic, designing intermediate query representations that align with question structures requires nontrivial insights into correspondence between question and query structures. \ours{} learns the correspondence end-to-end thanks to the grounding mechanism.

\subsection{Evaluation methodology}
\citep{keysers2019measuring} note that the structure of the development set is similar to the test set. Therefore, hyperparameter tuning could leak information from the test set. Unfortunately, not all baselines follow this evaluation procedure. Therefore, we report results obtained via development set tuning separately in Table~\ref{tab:on_dev}.
Results without tuning on development sets are not reported in \citet{guo2020hierarchical}, \citet{herzig2021unlocking}, and \citet{das2021case}.
All results in Table~\ref{tab:not_on_dev} are obtained without tuning on development sets.

\subsection{Ablations}

We benchmarked the following ablations of \ours{}:
\begin{itemize}
    \item \textit{Graph decoding only}~~This is the bare graph decoder architecture described in Section~\ref{sec:seq2gr}, which uses a sequence encoder without awareness of syntactic composition and does not have the grounding mechanism.
    \item \textit{Syntax-aware graph decoding}~~This is the architecture described in Section~\ref{sec:cseq2gr}, with a graph decoder on top of a sequence encoder aware of the ``A and B" syntax.
\end{itemize}

The purpose of the ablation study is to validate the effectiveness of both (1) syntax-aware sequence encoders and (2) the grounding mechanism.
We also compare graph decoding with poset decoding to show the benefit of graph decoding.

\subsection{Results and discussion}

We report performance of models tuned with and without development sets in Table~\ref{tab:on_dev} and Table~\ref{tab:not_on_dev} respectively.
Remarkably, in both settings, on the MCD1 split \ours{} attains accuracies attained previously only with random splits, effectively solving the split.

On the MCD2 and MCD3 splits of the CFQ dataset, \ours{} consistently outperforms the pre-trained T5 models regardless of their size (without the fine-tuning trick proposed by \citet{herzig2021unlocking}), as well as HPD and CBR-KGQA.
T5 performs remarkably when fine-tuned with lossy and reversible intermediate representations (LIR + RIR), though the results are not very surprising given that the LIRs and RIRs used for fine-tuning are tailored specifically to the CFQ dataset and can thus drastically ease the task.

Our error analysis shows that \ours{} is not able to completely solve the remaining two splits (MCD2 and MCD3) primarily as it only incorporates one type of syntactic composition, namely conjunctive queries of the form of ``A and B".
Pre-trained approaches such as T5 support a wider range of syntactic structures due to a more general training objective. However, rich syntactic structures acquired from pre-training contribute little to the compositional generalization.

We find that performance of T5 models are close to our graph decoder without any added compositionality (see Table~\ref{tab:not_on_dev}).
This suggests that pre-trained language models understand the syntax of conjunctive queries, possibly because their pre-training corpus contains conjunctive queries.
However, their generally low performance indicates that they do not utilize the information. Domain specific IRs could mitigate these challenges but are complex to apply to a real-world dataset like CFQ.

The ablation study verifies that graph decoder, syntax-aware sequence encoder, and the grounding mechanism are all important to compositional generalization. Grounding with graph decoding alone (no permutation invariance) results in a state-of-the-art model on the MCD1 split. However, this model struggles with MCD2 and MCD3. We also find that a directed graph is a better representation of conjunctive queries than poset decoding. This is due partially to the richer structure of a graph which is a natural fit for common query patterns.

\subsection{Future work}
There are several clear areas of future work beyond \ours{}. First, our method is highly complementary with pre-training based approaches to improving compositional generalization. Second, incorporating more complex syntactic structures which will likely further boost our results. Error analysis revealed conjunctive structures predominantly help improve accuracy in the MCD1 split but have less impact on the other splits.

\section{Related work}\label{sec:related}
Various approaches to the compositional generalization challenge posted by the CFQ dataset has been explored in prior or contemporary works, including \citet{guo2020hierarchical}, \citet{das2021case}, and \citet{herzig2021unlocking}.
These approaches are discussed in more detail in Section~\ref{sec:experiments}.

Another promising approach that has received relatively less attention so far is grammar induction, which can potentially derive grammar rules directly from question-query pairs.
Grammar induction methods, such as \citet{zettlemoyer2012learning}, typically assumes a limited set of grammar rules to bootstrap the model, and then search some grammar spaces to find grammars that can lead to successful parsing of observed questions.

The idea of grammar induction has inspired various work that to different extent solved the SCAN dataset, such as \citet{nye2020scanneuralprogramsynthesis} or \citet{ chen2020compositional}.
The advantage of grammar induction methods is that they can potentially identify the complete set of transformation rules and thus attain perfect compositional generalization.
However, grammar induction methods are generally search-based, which limits their scalability to long sentences due to the size of search spaces.

Additionally, there has been considerable research in the semantic parsing literature to design neural network architectures that incorporate different query structures, including tree \citep{dong2016language}, graph \citep{buys2017robust, damonte2016incremental, lyu2018amr, fancellu2019semantic}.
However, these architectures only incorporating query structures without incorporating syntactic structures in questions.
Our ablation study (Table~\ref{tab:ablations}) indicates that only incorporating query structure is insufficient for compositional generalization.
Our graph decoder alone only attains performance on bar with T5 models.

Similar to our work, \citet{russin2019compositional} also proposes to improve the compositional generalization of seq2seq models using attention.
However, their work only studies token-level attention without consideration of syntactic or semantic structures.
Both \citet{russin2019compositional} and \citet{Gordon2020Permutation} use part-of-speech (PoS) tags to attain some level of invariance among words that share the same PoS.

Finally, in the domain of semantic parsing, prior to \citet{keysers2019measuring}, \citet{finegan2018improving} proposed to split datasets such that training and test sets contain no common SQL patterns.
Although this approach increases task difficulty, different SQL query patterns may still share similar substructures, which enables neural networks to solve the tasks relatively easily using the ``mix-and-match" strategy \citep{lake2018generalization}.

\section{Conclusion}
In this paper we propose \ours{} to make compositionally generalizable predictions of conjunctive query from natural language questions.
Our model consists of a graph decoder that captures permutation invariance in conjunctive queries, a sequence encoder that is aware of syntactic composition, and an attention mechanism that enables strong association between syntactic and semantic compositions.
The proposed method solves the MCD1 split of the challenging CFQ dataset, and improves the state-of-the-art of the other two splits. Notably, \ours{} significantly outperforms competitive baselines including large pre-trained models (such as T5) as well as domain-specific models. Careful ablations of our method demonstrate the importance of both graph decoding and grounding.

\section*{Acknowledgments}
We thank Lisa Dunlap, Daniel Furrer, Ajay Jain, Daniel Rothchild, Nathan Scales, Rishabh Singh, Justin Wong and Marc van Zee for their help. In addition to NSF CISE Expeditions Award CCF-1730628, this research is supported by gifts from Amazon Web Services, Ant Financial, Ericsson, Facebook, Futurewei, Google, Intel, Microsoft, NVIDIA, Scotiabank, Splunk and VMware.

\clearpage
\bibliography{custom}
\bibliographystyle{acl_natbib}
\balance{}

\end{document}